\documentclass[letterpaper,journal]{IEEEtran}
\usepackage{amsmath,amsfonts}
\usepackage{algorithmic}
\usepackage{algorithm}
\usepackage{array}
\usepackage[caption=false,font=normalsize,labelfont=sf,textfont=sf]{subfig}
\usepackage{textcomp}
\usepackage{stfloats}
\usepackage{url}
\usepackage{verbatim}
\usepackage{graphicx}
\usepackage{orcidlink}
\usepackage{cite}
\usepackage{lipsum}
\usepackage{booktabs}
\begin{document}
\title{Unlocking Stopped-Rotor Flight: Development and Validation of SPERO, a Novel UAV Platform}

\author{Kristan Hilby\,\orcidlink{0000-0002-1562-3759} and Ian Hunter\,\orcidlink{0000-0002-2218-9516} 
}



\maketitle

\begin{abstract}
Stop-rotor aircraft have long been proposed as the ideal vertical
takeoff and landing (VTOL) aircraft for missions with equal time spent
in both flight regimes, such as agricultural monitoring, search and
rescue, and last-mile delivery. Featuring a central lifting surface
that rotates in VTOL to generate vertical thrust and locks in forward
flight to generate passive lift, the stop-rotor offers the potential
for high efficiency across both modes. However, practical
implementation has remained infeasible due to aerodynamic and
stability conflicts between flight modes. In this work, we present
SPERO (Stopped-Penta Rotor), a stop-rotor uncrewed aerial vehicle (UAV) featuring a flipping and latching wing, an active center of pressure
mechanism, thrust vectored counterbalances, a five-rotor architecture,
and an eleven-state machine flight controller coordinating
geometric and controller reconfiguration. Furthermore, SPERO
establishes a generalizable design and control framework for
stopped-rotor UAVs. Together, these innovations overcome longstanding
challenges in
stop-rotor flight and enable the first stable, bidirectional
transition between VTOL and forward flight.
\end{abstract}


\section{Introduction}
\IEEEPARstart{V}{ertical} take-off and landing (VTOL) uncrewed aerial vehicles (UAVs) have enabled the presence of aerial robotics in remote and undeveloped environments, performing tasks often deemed too dangerous or mundane for humans \cite{Lyu2023UnmannedSurvey,Watts2010SmallSurveys, Ozdemir2014DesignSystem} through enhanced safety, versatility, and agility \cite{Zhou2020AnVehicles}. Existing VTOL configurations--including multi-rotors \cite{Johnson1994HelicopterTheory, Quan2017IntroductionControl}, tiltrotor \cite{Thomason1990TheLearned, Gertler2011V-22Congress, Lopez2021BellData, Fonte2019EnhancedTiltrotor, Maisel2000TheFlight, Conner2002NASA/Army/BellProgram}, tiltwing \cite{Ransone2002AnContributions, Nichols1990TheLearned, Konrad1967FlightTransport, Pegg1962SummaryAircraft, Mitchell1963Full-ScalePhenomena}, and tailsitter \cite{Tesla1928ApparatusTransportation, Anderson1983AnDevelopment, Knoebel2006PreliminaryTailsitters, Taylor2003SkyTote:System, Ito2019ConceptVehicles}--are effective but often compromise efficiency in hover or forward flight \cite{Finger2017AAircraft, Welch2010AssessmentApplication, Bacchini2019ElectricComparison}. As missions evolve from short point-to-point flights toward complex, long-duration operations, efficiency across both regimes becomes critical \cite{Hilby2025DesignWings}.

\begin{figure}[!t]
    \centering
    \includegraphics[width=\linewidth]{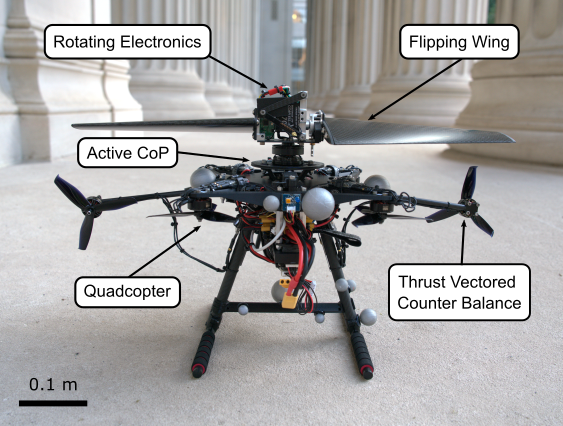}
    \caption{SPERO, a novel stopped-rotor UAV capable of stable bidirectional transition flight enabled by the highlighted aerodynamic and control features.}
    \label{fig:mainimage}
\end{figure}

One proposed configuration to address this trade-off is the stop-rotor aircraft, which employs a central lifting surface that rotates to provide vertical thrust in VTOL and locks in place to act as a fixed wing in forward flight. This dual functionality offers potential for greater speed, range, and payload capacity than conventional VTOL UAVs. However, despite significant research on stop-rotor concepts \cite{Herrick1938SomeAirplane, Herrick1951MultipleAircraft,Smith1968HotAircraft,Guertin1985DevelopmentRSRA/X-Wing,Lane1987TheReport, Linden1986RSRA/X-WingReport, Sikorsky2007TheLegacy,Jenkins2003AmericanEdition,Mitchell2003TheFly,Gai2011ModelingVehicle, Gao2019TrimMode, Gao2022AModel, Wang2023AdaptiveUncertainties, Zhu2023AdaptiveAircraft, Zhang2023FlightMode, Gao2022FlightLanding, Tayman1998AnConcept, Tayman2011Stop-rotorAircraft, HE2022FullUAV, Vargas-Clara2012DynamicsVehicle,Low2017DesignTHOR, Ramamurti2005ComputationalConfigurations, Kellogg2003DevelopmentConfigurations,Bevirt2018AerodynamicallyRotors, Sinha2015DesignUAV, Brown2023TheIdea}, to the best of our knowledge, experimental demonstration of stable and reversible bidirectional transitions remains unreported.

Stable stopped-rotor flight faces challenges across four domains. First, aerodynamic requirements across the wings changes as a function of flight mode \cite{Hilby2023DesignMechanism, Hilby2025Slat-InspiredVehicles}. Second, the geometric relationship between center of pressure (CoP) and center of gravity to ensure aerodynamic stability changes across flight modes. Third, the additional forces needed for stability and operation across flight modes differs. Finally, the rotor experiences a temporary loss of lift during transition, complicating altitude control.
\IEEEpubidadjcol

This work presents SPERO, a stopped-penta rotor UAV capable of stable bidirectional flight, shown in Fig. \ref{fig:mainimage}. SPERO integrates a flipping wing enabled by separated rotor and body avionics to manage airfoil directionality, an active center of pressure control mechanism for stability across flight regimes, and a quadcopter-based architecture to mitigate temporary loss of lift. Further, insights derived from theoretical stability analysis and the development of a mode-dependent state-machine controller to coordinate transitions. Building upon theoretical stability analysis, we further develop a mode-dependent state-machine control strategy that coordinates these features to enable smooth and robust transitions between flight modes. We contribute a comprehensive design and control methodology for stop-rotor UAVs, bridging aerodynamic requirements, vehicle geometry, and controller synthesis into a cohesive framework. We validate the proposed approach through hardware implementation and real-world flight testing, demonstrating SPERO’s ability to achieve stable and reversible bidirectional transitions. 

The present work establishes the foundational design methodology, modeling framework, and control architecture for stopped-rotor UAVs, upon which subsequent validation and model refinement efforts are built. These key contributions are summarized as follows:
\begin{enumerate}
    \item A unified design methodology that establishes the aerodynamic, geometric, and control requirements necessary for achieving stable operation of stop-rotor UAVs.
    \item A mode-dependent control framework that integrates state-machine logic with stability-driven control strategies to manage transitions across flight regimes.
    \item The first experimental demonstration of stable, reversible, and bidirectional transitions in a stop-rotor UAV, validated through real-world flight testing. 
\end{enumerate}

\begin{figure*}[!b]
    \centering
    \includegraphics[width=0.8\linewidth]{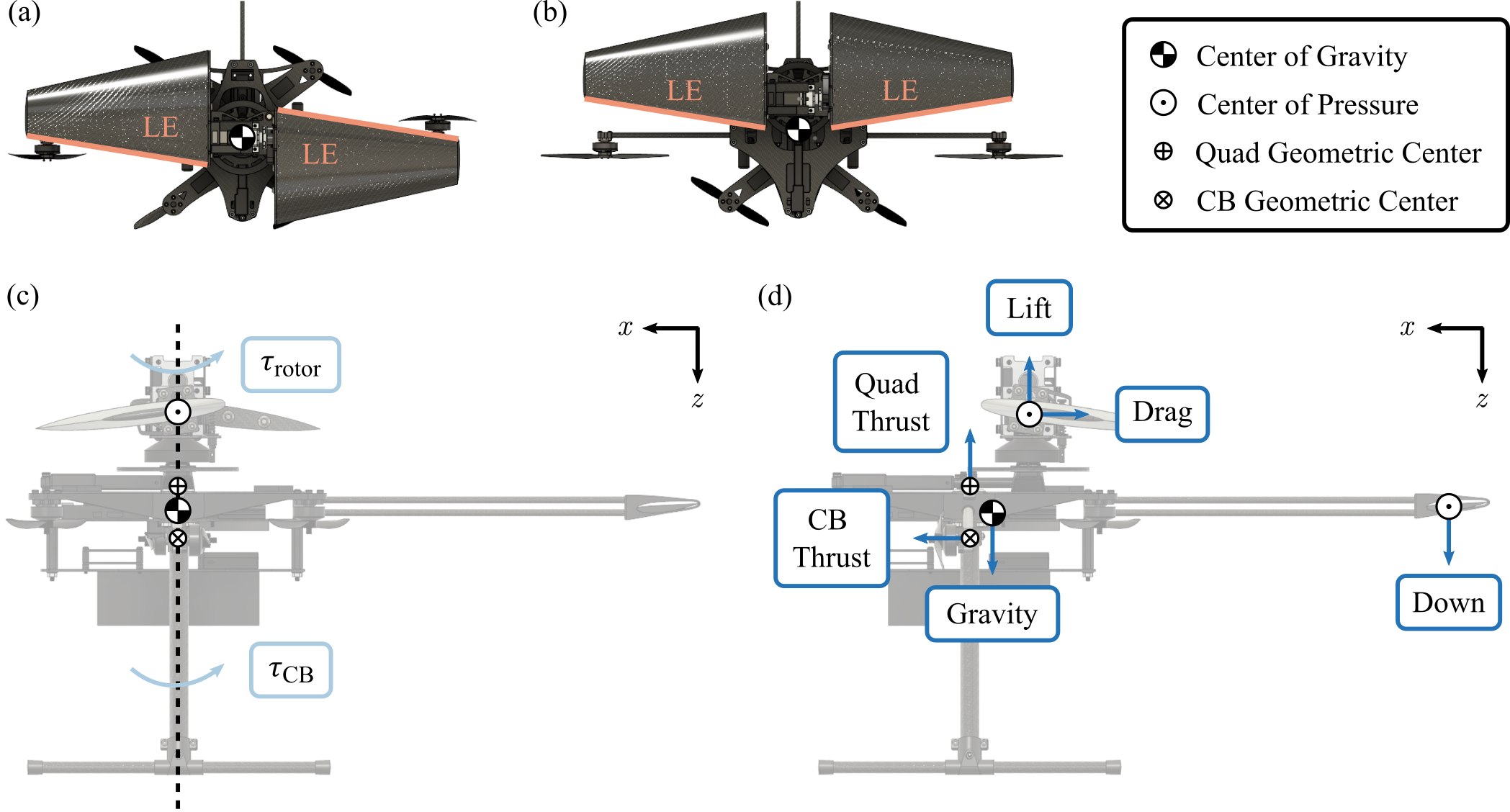}
    \caption{Overview of aerodynamic and geometric requirements across flight modes. (a) In VTOL, the preferred airfoil orientation places the leading edges on opposite sides of the wing (orange lines) to ensure balanced thrust during hover. (b) In forward flight, the leading edges are aligned on the same side to maximize lift and minimize drag. (c) During VTOL, gyroscopic effects are minimized by aligning the axis of rotation (i.e., effective wing center of pressure) with the geometric center of counterbalance torque ($\tau_{\mathrm{CB}}$). (d) In forward flight, placing the center of pressure aft of the center of gravity promotes passive pitch stability by generating a restoring moment under aerodynamic disturbances.}
    \label{fig:aeroreq}
\end{figure*}

\section{Feasibility Conditions for Stop-Rotor Flight}\label{sec:desprereq}
Achieving stable and efficient flight across VTOL, forward flight, and transition regimes imposes a set of conflicting aerodynamic, geometric, and control constraints for stop-rotor aircraft. These constraints, summarized in Fig. \ref{fig:aeroreq}, fundamentally shape the design of lifting surfaces, vehicle layout, and control architectures, and directly motivate the key design features implemented in SPERO. Through this work, we refer to the set of conflicting requirements as feasibility conditions (FCs), as they are necessary for the vehicle to operate effectively across VTOL, forward flight, and transitions. 

\subsection{Aerodynamic Requirements (FC1)}\label{sec:airfoilreq}
Efficient operation in VTOL and forward flight requires different wing configurations optimized for distinct aerodynamic objectives. In VTOL, wings with moderate twist balance the lift distribution and symmetric airfoils with the leading edges on opposite sides of the wing ensure equal thrust across the wing is generated during hover (Fig. \ref{fig:aeroreq}a).

Conversely, in forward flight, the aerodynamic efficiency (i.e., the ratio between lift and drag) dominates: wings with cambered airfoils, moderate taper, and leading edges aligned on the same side of the wing are preferred (Fig. \ref{fig:aeroreq}b). While camber, taper, and twist influence the optimal wing design, this work focuses exclusively on airfoil directionality requirements, leaving broader aerodynamic optimization to future studies.   

\subsection{Geometric Layout for Stability (FC2)}\label{sec:geomreq}
Differences in force generation across flight modes impose unique geometric constraints on the vehicle layout to ensure stability. In VTOL, the rotating wing introduces two key considerations. First, because the wing spins about the yaw axis, the vehicle must maintain sufficient yaw control authority to stabilize heading. Second, misalignment between the axis of rotation of the wing and the vehicle’s center of gravity possibly introduces unwanted gyroscopic precession. To mitigate these effects, the rotational axis of the wing and any additional counter-yaw torque should be coaxially aligned with the center of gravity, as illustrated in Fig. \ref{fig:aeroreq}c. 

By contrast, in forward flight, pitch stability is the primary consideration as constant lift and drag forces can generate moments about the center of gravity of the vehicle. For passive pitch stability, it is desirable for the center of pressure to be aft the center of gravity (Fig. \ref{fig:aeroreq}d), such that aerodynamic disturbances induce a restoring moment. 

\subsection{Auxiliary Forces (FC3)}\label{sec:auxforces}
Across flight modes, different control forces are needed to ensure stable operation. In VTOL, as briefly discussed in Section \ref{sec:geomreq}, a counter-torque is needed on the body to compensate for rotor acceleration and drag. This additional control torque is depicted as $\tau_{\mathrm{CB}}$ in Fig. \ref{fig:aeroreq}c. 

In forward flight, a thrust in the $x$ direction is needed to accelerate the vehicle to trim speed in transition, as well as compensate for drag and maintain vehicle speed once trimmed conditions are met. This additional forward thrust is denoted as CB thrust in Fig. \ref{fig:aeroreq}d.

\begin{figure*}[!b]
    \centering
    \includegraphics[width=\linewidth]{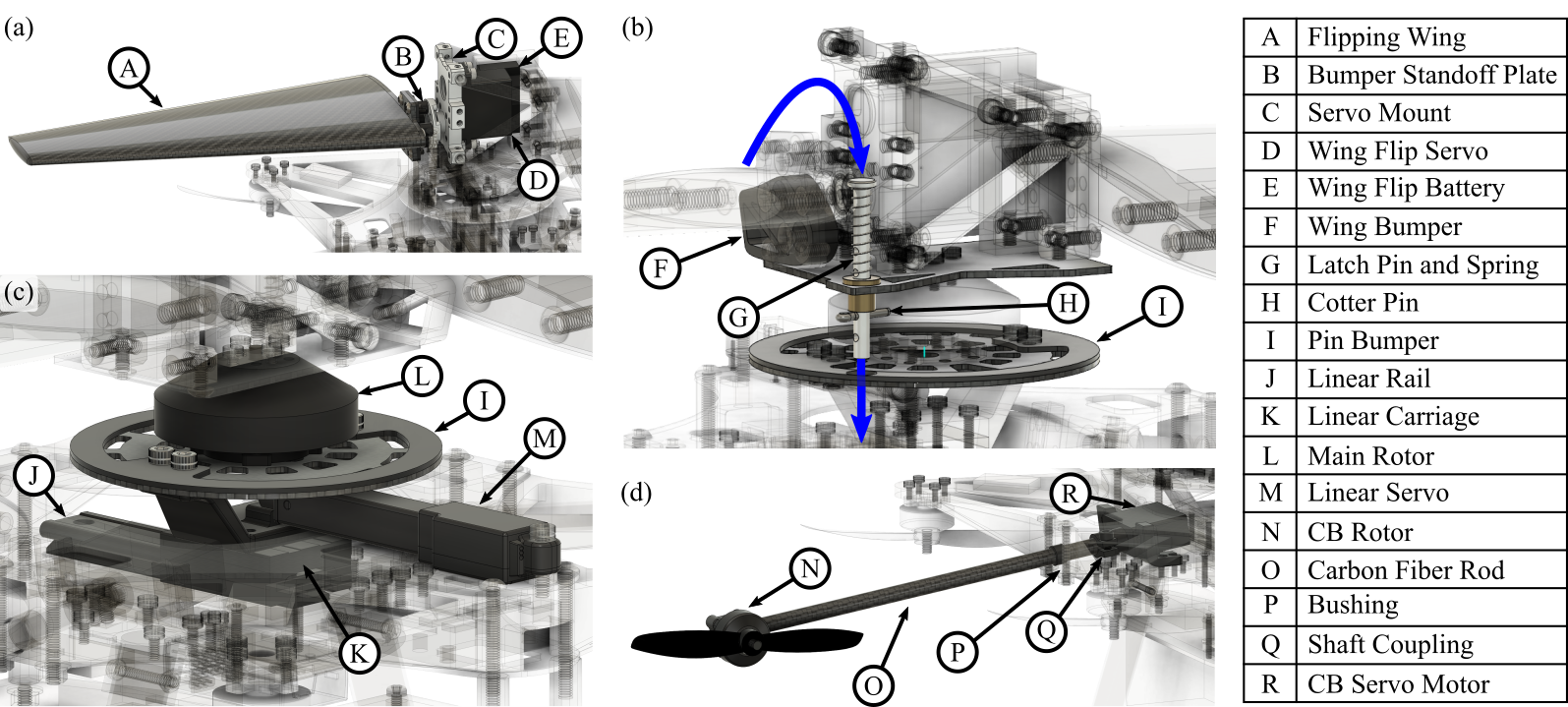}
    \caption{Implementation of SPERO’s key features. (a) Flipping wing mechanism driven by a servo and supported by a mount and plate. (b) Wing latching system comprising a spring-loaded latch pin, cotter pin, and bumper that secure the wing in the forward-flight configuration; the blue arrow illustrates the bumper trajectory during flipping. (c) Active center of pressure mechanism, where a linear servo drives a carriage along a rail to reposition the main rotor relative to the vehicle body, enabling passive stability across flight modes. (d) Counterbalance system with a servo-actuated shaft and coupling, providing adjustable control forces to accommodate the differing requirements of VTOL and forward flight. All components are indexed in the accompanying table.}
    \label{fig:detaileddesign}
\end{figure*}

\subsection{Lift Continuity (FC4)}\label{sec:lossoflift}
The working principle of stop-rotor UAVs relies on a single lifting surface operating in two distinct modes: In VTOL, lift is generated through rotor-induced circulation, whereas in forward flight, lift arises from the vehicle's translational motion. Because the lift generated by the wing scales with the square of effective airspeed ($L \propto v^2$), transition periods inherently produce a temporary loss of lift as the flow conditions change. 

During VTOL to forward flight (i.e., forward transition), the rotor decelerates and eventually locks to function as a fixed wing. As the rotor slows, the relative airspeed across the wing diminishes, causing the lift contribution to reduce to zero. Similarly, during forward flight to VTOL (i.e., backward transition), the freestream velocity diminishes and the wing again produces negligible lift.

In both cases, the temporary loss of wing lift removes altitude control authority and can destabilize the vehicle unless special consideration is taken. 

\section{Key Design Features} \label{sec:design}
In light of the requirements outlined in Section \ref{sec:desprereq}, we propose SPERO (Fig. \ref{fig:mainimage}), a 2.7 kg stopped-penta rotor UAV, featuring five key design features (DFs) that collectively address the identified challenges: (DF1) a flipping and precision latching wing mechanism to dynamically reorient the airfoil direction across flight modes and secure the wing in forward flight; (DF2) an active center of pressure positioning mechanism to manage the differing geometric stability requirements; (DF3) a quadcopter platform to maintain altitude control during transitions; (DF4) thrust-vectored counterbalances to provide yaw and thrust authority in VTOL and forward flight, respectively; and (DF5) a distributed avionics architecture, comprising independent top-rotor and main body systems that coordinates the actuation and control of the preceding design features while also managing flight control integration, communication, and telemetry. The mapping between feasibility conditions and design features are summarized in Table \ref{tab:dp_df_mapping}.

\begin{table}[!t]
\centering
\caption{Mapping Between Feasibility Conditions (FC) and Key Design Features (DF)}
\label{tab:dp_df_mapping}
\renewcommand{\arraystretch}{1.25} 
\setlength{\tabcolsep}{5pt}       
\begin{tabular}{lccccc}
\hline
 & \textbf{DF1} & \textbf{DF2} & \textbf{DF3} & \textbf{DF4} & \textbf{DF5} \\ 
  & (\S\ref{sec:d_flipwing}) &  (\S\ref{sec:d_activecop}) & (\S\ref{sec:d_thrustveccbs}) & (\S\ref{sec:d_quad}) & (\S\ref{sec:d_avionics})\\ 
\hline
\textbf{FC1} (\S\ref{sec:airfoilreq}) & \checkmark &             &             &            & $\circ$\\ 
\textbf{FC2} (\S\ref{sec:geomreq})    &            & \checkmark  &             &            & $\circ$\\ 
\textbf{FC3} (\S\ref{sec:auxforces})  &            &             & \checkmark  & $\circ$    & $\circ$\\ 
\textbf{FC4} (\S\ref{sec:lossoflift}) &            &             & $\circ$     & \checkmark & $\circ$\\ 
\hline
\end{tabular}
\\[6pt]
\begin{minipage}{0.9\linewidth}
\footnotesize \textit{Legend:} \checkmark = primary contribution (directly addresses the requirement); $\circ$ = secondary contribution (indirectly supports the requirement, but is not the main enabler; \S~= Section.
\end{minipage}
\end{table}

\begin{figure*}[!b]
    \centering
    \includegraphics[width=\textwidth]{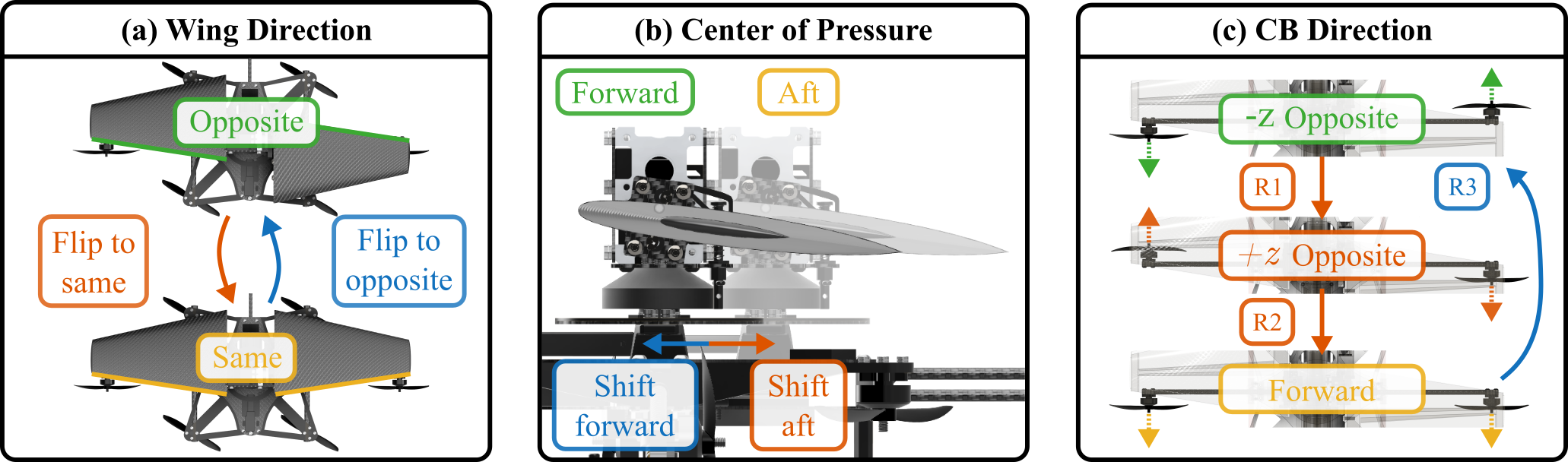}
    \caption{Overview of the geometric transformations undergone by SPERO during bidirectional flight. Color coding is consistent across panels: VTOL configurations, forward transition configurations, backward transition configurations, and forward flight configurations are depicted in green, orange, blue, and yellow, respectively. (a) Wing direction configurations, showing the leading edges oriented on opposite sides during VTOL flight and aligned on the same side during forward flight. (b) Center of pressure configurations used to adjust the vehicle’s mass distribution. (c) Counterbalance (CB) directions used to adjust net forces on the vehicle. R1 – R3 denote transitions between discrete “-z”, “+z”, and “forward” configurations.}
    \label{fig:controlmodes}
\end{figure*}

\begin{figure*}[!t]
    \centering
    \includegraphics[width=\linewidth]{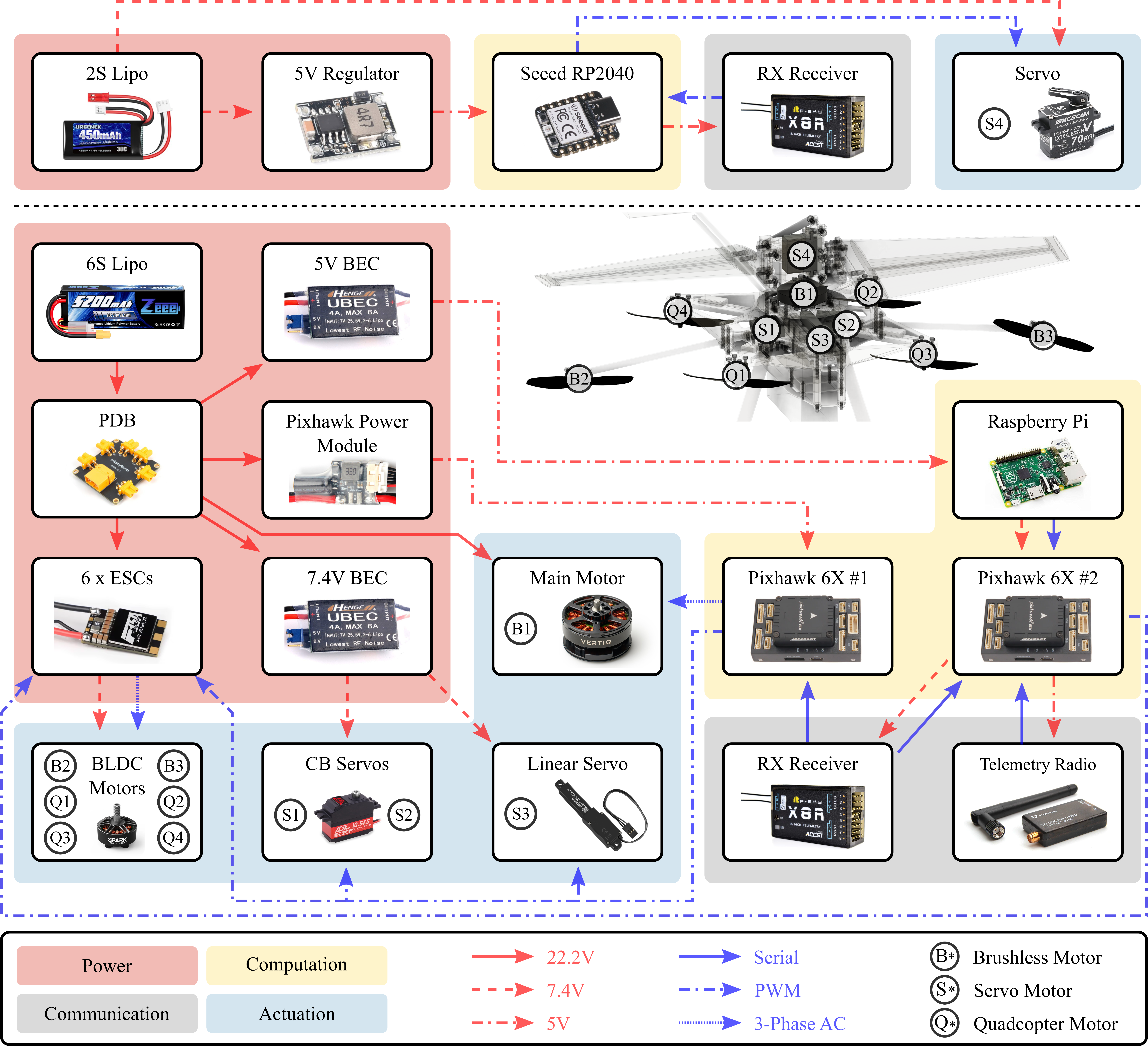}
    \caption{Avionics architecture of SPERO, showing the power (red), computation (yellow), communication (gray), and actuation (blue) components. Power signals are indicated by red lines, with solid, dashed, and dot-dashed styles corresponding to 22.2 V, 7.4 V, and 5 V, respectively. Communication signals are shown in blue, where solid, dot-dashed, and dashed lines represent serial, PWM, and three-phase AC sinusoidal signals, respectively. The positions of the actuation components are overlaid on SPERO, with indices \textbf{B}, \textbf{S}, and \textbf{Q} denoting brushless motors, servo motors, and quadcopter motors, respectively. Elements above the dashed black line belong to the top-rotor avionics, while those below belong to the main body avionics.}
    \label{fig:avionics}
\end{figure*}

The mechanical implementation of DF1 -- DF4 is provided in Fig. \ref{fig:detaileddesign}, while Fig. \ref{fig:controlmodes} summarizes the possible vehicle configurations resulting from these features. The configuration labels defined in Fig. \ref{fig:controlmodes} are used consistently in subsequent sections. The avionics relating to DF5 are summarized in Fig. \ref{fig:avionics}. Table \ref{table:designparams} summarizes the relevant design parameters, including key dimensions, mass properties, and transition properties. Finally, all design files, including CAD models, 3D-printable files, machining drawings, and an interactive 3D viewer, are available online \cite{Redacted2025SPERODocumentation}. The following sections examine each design feature in detail, discussing the relation to feasibility conditions and mechanical design.

\begin{table}[!t]
\centering
\caption{Physical Parameters for SPERO. Summary of the key dimensions, mass properties, and performance values for SPERO. CoP refers to the center of pressure, and CG refers to the center of gravity. 
*Reference Appendix \ref{app:copcalc} for calculation
}
\label{table:designparams}
\begin{tabular}{lcc}
\hline
\textbf{Parameter} & \textbf{Value} & \textbf{Unit} \\
\hline
\multicolumn{3}{l}{\textbf{Dimensions}} \\
Body bounding box length ($l$) & 0.665 & m \\
Body bounding box width ($w$) & 0.619 & m \\
Body bounding box height ($h$) & 0.350 & m \\
Central rotor diameter & 0.588 & m \\
Single wing span & 0.250 & m \\
Quadrotor and counterbalance diameter & 0.127 & m \\
\hline
\multicolumn{3}{l}{\textbf{Mass Properties}} \\
Total mass & 2.727 & kg \\
Single wing mass & 0.168 & kg \\
Single counterbalance mass & 0.086 & kg \\
Wing flip hardware mass & 0.145 & kg \\
Passive wing latch hardware mass & 0.011 & kg \\
Active center of pressure mass & 0.350 & kg \\
Main battery mass & 0.460 & kg \\
Top rotor battery mass & 0.040 & kg \\
\hline
\multicolumn{3}{l}{\textbf{Transition Properties}} \\
Measured forward transition time & 4.2 & s \\
Measured backward transition time & 3.8 & s \\
Wing flip angle & 190 & deg \\
Maximum forward flight CoP to CG distance* & 0.03 & m \\
\hline
\end{tabular}
\end{table}

\subsection{Flipping and latching wing ensures ideal aerodynamic configuration (DF1)}\label{sec:d_flipwing}
As outlined in Sec. \ref{sec:airfoilreq}, achieving stable and efficient flight requires the airfoil direction of the wings to adapt between VTOL and forward flight to satisfy distinct aerodynamic requirements. Prior work has explored reversible morphing wings to achieve this adaptation \cite{Hilby2023DesignMechanism, Hilby2025Slat-InspiredVehicles, Hilby2025DesignVehicles, Niemiec2014ReversibleFlight, Niemiec2015Leading-andRotors}, but these solutions are often mechanically complex and poorly suited for lightweight UAV platforms.

SPERO addresses this challenge with a servo-driven flipping mechanism. As shown in Fig. \ref{fig:detaileddesign}a, one of the wings is mounted to a servomotor through a series of support brackets, allowing the wing to rotate about the axis along the span. By actuating the servo, the wing transitions between the \textit{Same} and \textit{Opposite} configurations via \textit{Flip to same} and \textit{Flip to opposite} transitions, as summarized in Fig. \ref{fig:controlmodes}a. The wing uses a symmetric NACA 0012 airfoil to ensure the same wing performance across rotations. 

To complement the flipping mechanism, a passive latching system (Fig. \ref{fig:detaileddesign}b) secures the wing in its forward-flight orientation, eliminating oscillations and ensuring rigidity under aerodynamic loads. The latch consists of a spring-loaded pin, bumper, catch, and cotter pin, engaging automatically as the wing rotates into position. During the reverse transition back to VTOL, the spring retracts the pin, releasing the wing and allowing the rotor to spin freely.

\subsection{Active center of pressure provides passive and dynamic stability (DF2)}\label{sec:d_activecop}
To satisfy the varying center of pressure alignment requirements described in Sec. \ref{sec:geomreq}, SPERO employs an active center of pressure positioning mechanism, illustrated in Fig. \ref{fig:detaileddesign}c. The main rotor and wing assembly are mounted on a carriage that translates along a linear rail by a servo, effectively shifting the center of pressure during actuation. 

During flight, this mechanism allows SPERO to position the center of pressure in the \textit{Forward} and \textit{Aft} orientations via \textit{Shift forward} and \textit{Shift aft} transitions, as summarized in Fig. \ref{fig:controlmodes}b.  

\subsection{Thrust vectored counterbalances ensure adequate control authority without design overhead (DF3)}\label{sec:d_thrustveccbs}
To meet the varying control force requirements described in Section \ref{sec:auxforces}, SPERO employs a thrust-vectored counterbalance assembly on each side of the vehicle, shown in Fig. \ref{fig:detaileddesign}d. Each assembly consists of a brushless motor mounted to a carbon fiber rod via a custom adapter. The carbon fiber rod is rotated by a servo motor to alter the direction of thrust from the brushless motor. This mechanism primarily satisfies the auxiliary force requirement (FC3) by actively controlling the direction of the forces introduced by the attached brushless motors. 

In VTOL and rotor acceleration, the counterbalances are oriented in opposite directions to produce a net torque $\tau_\mathrm{CB}$ on the vehicle, corresponding to the \textit{-z Opposite} configuration in Fig. \ref{fig:controlmodes}c. During rotor deceleration, the counterbalances remain arranged in opposite directions, but the torque direction is reversed by changing the direction of both counterbalances. This configuration is referred to as the \textit{+z Opposite} orientation shown in Fig. \ref{fig:controlmodes}c. Finally, in forward flight, both counterbalances are aligned forward, producing thrust for acceleration and cruise, denoted as the \textit{Forward} configuration in Fig. \ref{fig:controlmodes}c. The transitions between these three configurations are denoted \textit{R1}, \textit{R2}, and \textit{R3} in Fig. \ref{fig:controlmodes}c.

Beyond their primary role of generating the necessary force and torque for stable operation, the counterbalances can also provide a secondary lift contribution during transitions. By orienting the counterbalances in intermediate angles between \textit{-z opposite} and \textit{+z opposite} configurations, the counterbalances augment lift during transition phases. This incidental contribution enhances lift continuity (FC4) but does not drive the mechanism’s design, making it a secondary effect rather than a primary requirement.

\subsection{Quadcopter architecture ensures stable altitude control through temporary loss of lift (DF4)}\label{sec:d_quad}
The temporary loss of lift (FC4) outlined in Section \ref{sec:lossoflift} is primarily addressed through the integration of a quadcopter architecture into the base of the vehicle, as annotated in Fig. \ref{fig:mainimage}. Four brushless motors are mounted directly to the base plate of the vehicle, providing supplemental thrust when the lift generated by the main rotor decreases. During transitions, the quadcopter motors gradually increase thrust to support the vehicle's weight, ensuring continuous lift and satisfying the primary requirement for lift continuity (FC4).

Beyond the primary contribution, the quadcopter motors also provide a secondary contribution to auxiliary force needs (FC3). In VTOL and rotor acceleration, the quadcopter motors introduce a small moment to assist the vehicle in yaw stabilization. Similarly, as the vehicle is accelerating to reach trim speed in forward flight, the quadcopter motors contribute a vertical force. This effect is considered secondary because the quadcopter's contributions are limited to minor adjustments, rather than increasing overall control authority. 

\subsection{Distributed avionics architecture enables mechanism actuation and flight control (DF5)}\label{sec:d_avionics}
SPERO employs the distributed avionics architecture shown in Fig. \ref{fig:avionics}. Unlike the other key design features (DF1 -- DF4), the avionics system (DF5) does not directly address any single fundamental necessity (FC1 -- FC4); instead, the avionics system, and specifically the distributed nature, enables the function of the other design features through power, computation, communication, and actuation modules.

Power modules regulate and distribute energy between SPERO's components. Computation modules perform state estimation and control allocation to drive the various components on SPERO. Communication modules relay data between the vehicle and a ground station. Finally, actuation modules contain the actuators of the vehicle that realizes DF1 -- DF4, shown in the inset image of Fig. \ref{fig:avionics}. All avionics components are available off the shelf.  

\subsubsection{Top-rotor avionics enable the flipping wing (DF1)}
The flipping wing mechanism (DF1) requires a dedicated set of electronics that are mounted on the rotating top-rotor assembly, shown above the dotted line in Fig. \ref{fig:avionics}. Because of the relatively high rotor speeds ($>$ 80 rad/s) and low vehicle weight ($<$ 3 kg), conventional slip-ring power transmission is impractical. Instead, a self-constrained top-rotor avionics package manages the wing flipping separate from the main body avionics.

Commands from a ground pilot are received by an RX module and converted into the appropriate rescaled pulse width modulated (PWM) signal to match the full servo range with a Seeed RP2040. The entire top-rotor avionics package is powered by a two-cell 2.96 Wh (400 mAh) battery via a 5V regulator. The relevant code for the Seeed RP2040 is available online \cite{Redacted2025ServoRescaling}. 

\subsubsection{Main body avionics enable stop-rotor processes, DF2 -- DF4 functionality, and general UAV operations}
The main body avionics of SPERO, shown below the dotted line in Fig. \ref{fig:avionics}, manages three responsibilities: 1) Enable stop-rotor processes, such as the main rotor integration and control, 2) coordination of DF2 -- DF4 functionality through the linear servo for active center of pressure positioning (DF2), the servos and brushless motors for the counterbalances (DF3), and the quadcopter system for lift continuity (DF4), and 3) executing general UAV operations such as flight control, state estimation, and communication. 

Computation is split between two Pixhawk 6X flight controllers. The first executes a custom stop-rotor control stack, managing the main motor, linear servo, and counterbalances, while the second runs the standard PX4 VTOL controller for quadcopter transitions. The separation of control loops enables decentralized coordination between the stop-rotor and quadcopter subsystems for improved robustness during testing. 

State estimation is provided via a Raspberry Pi 4 integrating motion-capture feedback. The custom PX4 build for the first Pixhawk is available online \cite{Redacted2025SPERODocumentation}. Communication with the ground station is handled via a telemetry radio for state streaming and an RX receiver for pilot commands.

\section{Simplified Dynamics, Stability Characterization, and Control Insight}\label{sec:modeling}
Building on the design insights established in the preceding sections, this section develops a simplified analytical model of SPERO's dynamics to characterize operational constraints and inform controller synthesis. We focus on yaw and altitude degrees of freedom, as these are most directly influenced by the rotor aerodynamics and inertia compared to roll and pitch. 

Specifically, we first derive simplified expressions for the yaw and altitude of the vehicle. Next, we develop a control policy informed by the derived dynamics to ensure stability. The stability of two closed-loop controllers are evaluated and the performance is compared. The section is concluded with a summary of the insights derived from the various analyses. All constants and variables used in this section are defined in the text and summarized in Table \ref{table:physicalparams}.  
\begin{table}[!t]
\centering
\caption{Variable definitions and aerodynamic parameters for modeling SPERO. CoP refers to the center of pressure.}
\label{table:physicalparams}
\begin{tabular}{l l c c}
\hline
\textbf{Symbol} & \textbf{Definition} & \textbf{Value} & \textbf{Unit} \\
\hline
\multicolumn{4}{l}{\textbf{Inertia Properties}} \\
$I_{\text{body}}$ & Body inertia about $z$-axis & 0.0345 & kg·m$^2$ \\
$I_{\text{rotor}}$ & Rotor inertia about $z$-axis & 0.0016 & kg·m$^2$ \\
\hline
\multicolumn{4}{l}{\textbf{Aerodynamic Properties}} \\
$\rho$ & Air density & 1.225 & kg/m$^3$ \\
$c_d$ & Wing drag coefficient & 0.05 & -- \\
$A_{\text{ref}}$ & Wing reference area & 0.056 & m$^2$ \\
$c_l$ & Wing lift coefficient & 0.87 & -- \\
\hline
\multicolumn{4}{l}{\textbf{Dynamic Variables}} \\
$\alpha_{\text{rotor}}(t)$ & Rotor yaw angle & -- & rad \\
$\alpha_{\text{body}}(t)$ & Body yaw angle & -- & rad \\
$\omega_{\text{rotor}}(t)$ & Rotor yaw rate & -- & rad/s \\
$\tau_u(t)$ & Yaw control torque & -- & N·m \\
$F_u(t)$ & Vertical control force & -- & N \\
$z(t)$ & Vehicle altitude & -- & m \\
$u_{\mathrm{yaw}}(t)$ & Yaw control input & -- & -- \\
$u_{\mathrm{alt}}(t)$ & Altitude control input & -- & -- \\
\hline
\multicolumn{4}{l}{\textbf{Geometric and System Properties}} \\
$r$ & Rotation axis to wing CoP distance & 0.10 & m \\
$m$ & Total vehicle mass & 2.727 & kg \\
$m_{\text{wing}}$ & Combined mass of both wings & 0.34 & kg \\
$m_{\text{rail}}$ & Mass of active rail mechanism & 0.51 & kg \\
\hline
\multicolumn{4}{l}{\textbf{Fundamental Values}} \\
$t$ & Time & -- & s \\
$s$ & Laplace variable & -- & s$^{-1}$ \\
$g$ & Gravitational acceleration & 9.81 & m/s$^2$ \\
\hline
\end{tabular}
\end{table}

\subsection{Simplified Vehicle Dynamics}
\begin{figure}[!t]
    \centering
    \includegraphics[width=\columnwidth]{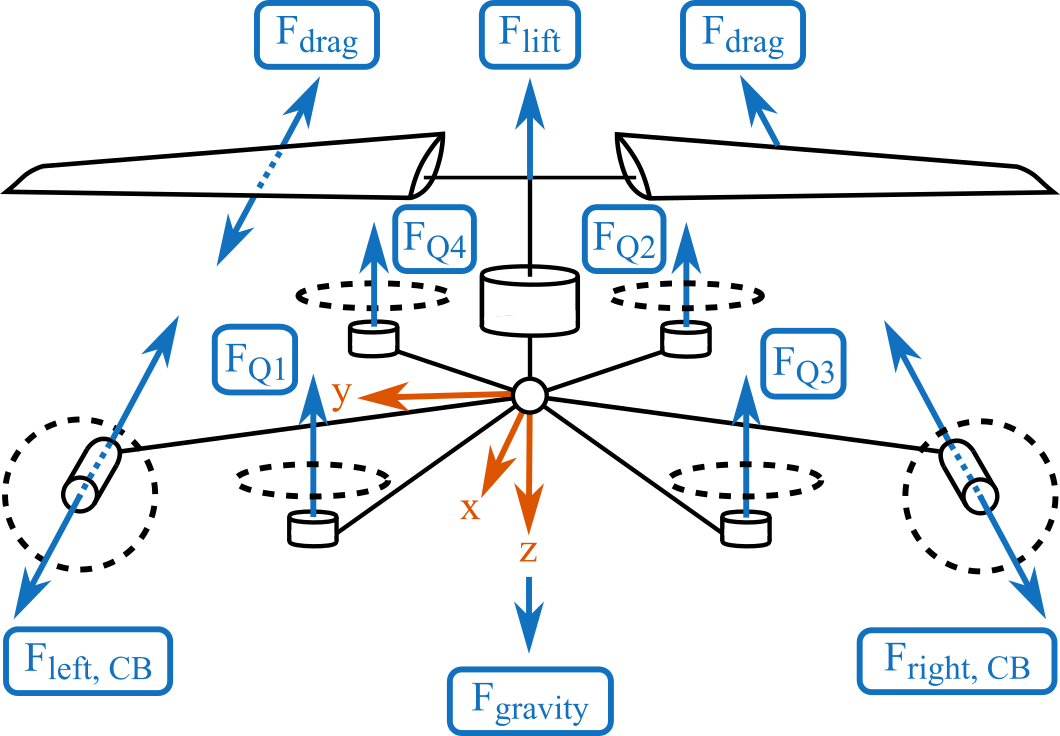}
    \caption{Simplified schematic illustrating key forces during operation. Blue arrows depict forces acting on the vehicle, including lift ($\mathrm{F}_\mathrm{lift}$), drag ($\mathrm{F}_\mathrm{drag}$), individual quadcopter thrust ($\mathrm{F}_\mathrm{Q1}$ -- $\mathrm{F}_\mathrm{Q4}$), individual counterbalance ($\mathrm{F}_\mathrm{left, CB}$, $\mathrm{F}_\mathrm{right, CB}$), and gravity ($\mathrm{F}_\mathrm{gravity}$). Dashed blue lines indicate the direction variability of forces across flight modes. Orange arrows define the vehicle axes, while dashed black lines indicate propeller boundaries.}
    \label{fig:reducedmodel}
\end{figure}

To derive tractable models suitable for supporting controller design and stability analyses, we develop simplified models of SPERO by applying Newton's second law to a simplified representation of SPERO, shown in Fig. \ref{fig:reducedmodel}. We aim to capture the dominant effects of rotor-driven dynamics while neglecting higher-order effects. Specifically, we model the yaw and altitude dynamics as these degrees of freedom are identified to be the most strongly coupled to rotor aerodynamics and inertia. Roll and pitch dynamics are ignored as they are decoupled from rotor effects in an ideal system and can be stabilized using well-studied quadcopter control methods. As such, the goal is not to capture the full six degree of freedom vehicle dynamics, but to derive tractable analytical models that provide insight into the stability and control across key flight regimes. 

\subsubsection{Modeling assumptions}\label{sec:modelingassumptions}
As the aim is to identify the key influence of rotor dynamics, the following assumptions are made: 
\begin{itemize}
    \item \textbf{Operational scope}: The analysis focus on VTOL and rotor spin up and down during transition as the yaw and altitude dynamics are most strongly coupled in these phases.
    \item \textbf{Force and torque modeling}: Forces and torques are assumed to be unidirectional and decoupled. Furthermore, only dominant forces and torques are considered: basic aerodynamic lift and drag from the main wing, torque from the counterbalance system, thrust from the quadcopter, and gravity.
    \item \textbf{Neglected aerodynamic effects}: Rotor inflow dynamics, unsteady vortex effects, and higher order unstable aerodynamic effects are neglected under the assumption of quasi-steady flow at a moderate Reynolds number,
    \item \textbf{Rigid body approximation}: All bodies are treated as rigid, thereby ignoring structural flexibility and aeroelastic effects. 
    \item \textbf{Actuator dynamics}: Actuator bandwidths are assumed to be fast relative to vehicle dynamics, allowing torque and thrust to be commanded instantaneously in the model.
    \item \textbf{Neglected sensor and communication delays}: Latency in sensing, computation, and communication are ignored.  
    \item \textbf{Lateral dynamics exclusion}: Roll and pitch dynamics are omitted as the preceding assumptions limit the effect of rotor dynamics. 
\end{itemize}

\subsubsection{Yaw dynamics}
Under the assumptions presented in the prior section, the yaw dynamics of SPERO during VTOL and transition flight are derived by applying Newton’s second law subject to the forces shown in Fig. \ref{fig:reducedmodel}, which yields the following expression:
\begin{equation}\label{eq:yawdynamics}
    I_{\text {body }} \ddot\alpha_{\text{body}}(t)=-I_{\text{rotor}} \cdot \ddot\alpha_{\text{rotor}}(t)-\frac{1}{2} \rho c_d A_{\text {ref}} r^3 \omega_{\text {rotor }}(t)^2+\tau_{\mathrm{u}}(t),
\end{equation}
where $I_{\mathrm{body}}$ and $I_{\mathrm{rotor}}$ are the moment of inertia of the body and rotor, respectively; $\alpha_{body}\left(t\right)$ and $\alpha_{rotor}\left(t\right)$ are the yaw angle of the body and rotor, respectively; $\rho$ is air density; $c_d$ is the drag coefficient of the wing; $A_{\mathrm{ref}}$ is the reference area of the wing; $\omega_{\mathrm{rotor}}\left(t\right)$ is the rotor speed; $r$ is the distance between the axis of rotation and the center of pressure; and $\tau_\mathrm{u}\left(t\right)$ is the control torque from the counterbalances.

\subsubsection{Altitude dynamics}
Similarly, applying the assumptions outlined in Section \ref{sec:modelingassumptions}, the altitude dynamics of SPERO during VTOL and transition flight are governed by the solution to Newton’s second law for forces along the z-axis of the body:
\begin{equation}\label{eq:altdynamics}
    m \cdot \ddot{z}(t)=m \cdot g-\frac{1}{2} \rho c_l A_{\text {ref }} r^2 \omega_{\text {rotor }}(t)^2-F_{\mathrm{u}}(t),
\end{equation}
where m is the total mass of the vehicle, $z\left(t\right)$ is the altitude, g is the gravitational constant, and $c_l$ is the lift coefficient of the wing.

\subsubsection{Notable modeling insights}\label{sec:modelinginsights}
Equation (\ref{eq:yawdynamics}) and Equation (\ref{eq:altdynamics}) reveal two important characteristics of the dynamics. First, both dynamics present a nonlinear dependence on rotor speed. Second, both present constant offset contributions from gravity, aerodynamic drag, and rotor acceleration. Both of these factors must be addressed to ensure accurate vehicle control and stabilization. These factors will be referenced again in the formulation of the control architecture in the following section.

\begin{figure*}[!b]
    \centering
    \includegraphics[width=0.9\textwidth]{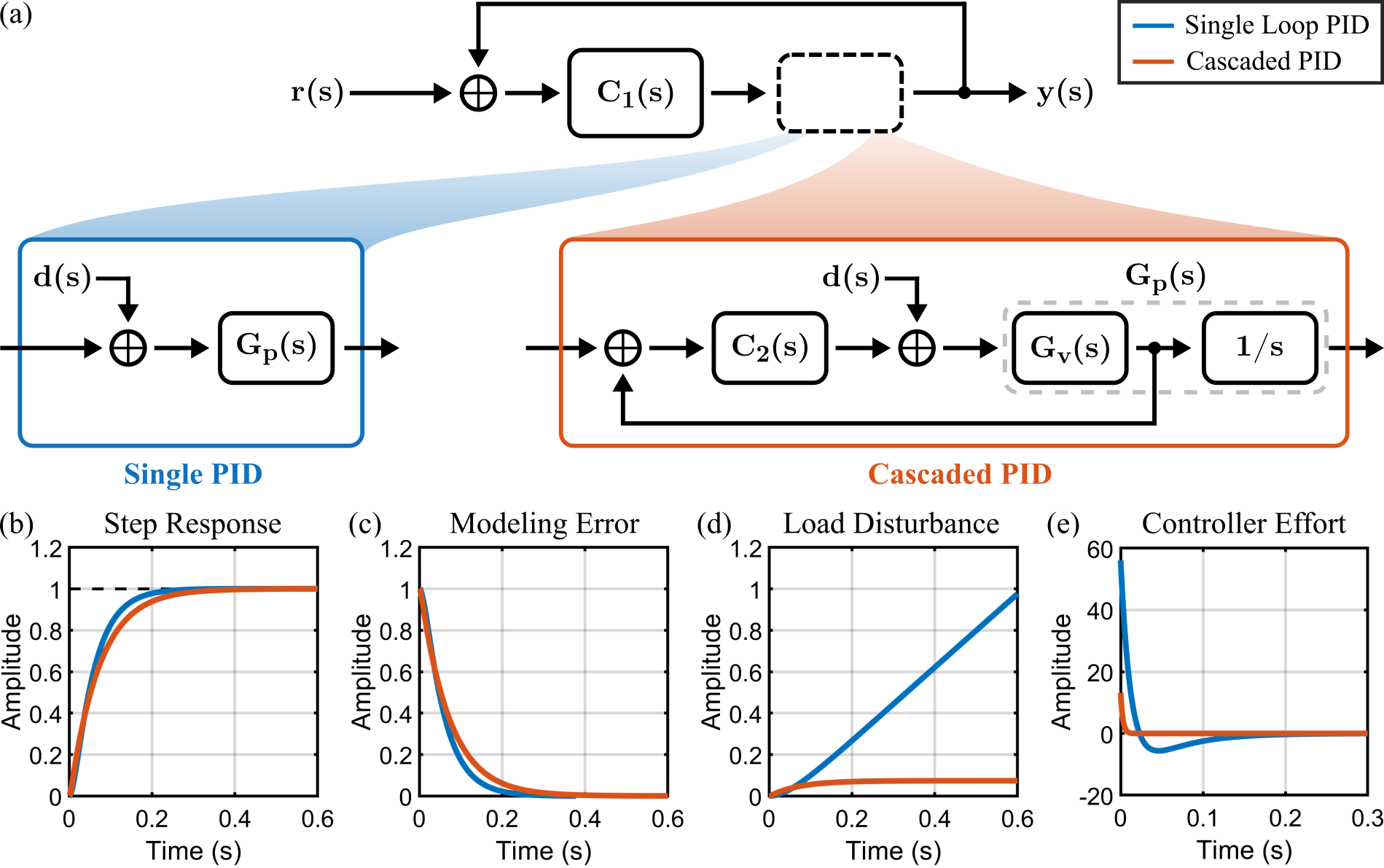}
    \caption{Comparison of single-loop and cascaded PID control architectures for SPERO’s yaw and altitude stabilization. (a) Control block diagram comparing the single loop (blue) and cascaded (orange) PID control structures. In the single-loop case, one PID controller $C_1\left(s\right)$ acts on the position error of the system, where the plant is modeled as $G_p\left(s\right)$. In the cascaded configuration, an outer-loop controller $C_1\left(s\right)$ regulates position while an inner-loop controller $ C_2\left(s\right)$ controls rate, based on the rate plant model $G_v\left(s\right)$. Simulation results using SPERO’s identified yaw dynamics highlight performance tradeoffs between the two architectures: (b) Step response shows comparable rise times, transient behavior, and steady-state accuracy. (c) When subject to modeling error, both loops reject the disturbance. (d) Under a step disturbance at the input to the plant, the cascaded system rejects the disturbance to a steady state value, while the single loop error continues to grow. (e) Controller effort required for tracking a step input showing that cascaded control exhibits lower peak demand than the single loop case.}
    \label{fig:controllercomps}
\end{figure*}

\subsection{Control Architecture}
SPERO's control architecture utilizes an augmented feedback control system, summarized in Fig. \ref{fig:controllercomps}, to stabilize the dynamics presented in Equation (\ref{eq:yawdynamics}) and Equation (\ref{eq:altdynamics}). Two feedback control structures are considered: single-loop PID and cascaded PI-PID. In the single-loop structure, a single controller, $C_1(s)$, acts on the error of position (i.e., yaw or altitude), shown by substituting in the path in the blue box in Fig. \ref{fig:controllercomps}a. In contrast, the cascaded structure separates control into two stages. First, an inner loop controls rate error (i.e., yaw rate or velocity in $z$) with $C_2(s)$. An outer controller, $C_1\left(s\right)$, controls the position error. The cascaded structure is shown by substituting the orange box into the path, as shown in Fig. \ref{fig:controllercomps}a.

The feedback control system is augmented in two ways. First, two forms of rotor speed are assumed. Second, feedforward linearization is used to compensate for the constant offset terms outlined in Section \ref{sec:modelinginsights}. Each augmentation is reviewed in further detail. 

\subsubsection{Nonlinear dependence on rotor speed}
In Case I, rotor speed is held constant such $\omega_{\mathrm{rotor}}\left(t\right)=\omega_{rotor}$ and $\ddot{\alpha}_{\mathrm{rotor}}\left(t\right)\ =\ 0$. As rotor speed cannot always be held constant (e.g., during transition), in Case II, rotor acceleration held constant (i.e., $\omega_{\mathrm{rotor}}\left(t\right)=\alpha_{rotor}t$).

\subsubsection{Constant offset terms}
Equation (\ref{eq:yawdynamics}) and Equation (\ref{eq:altdynamics}) present linear offset terms related to gravity, aerodynamic effects, and rotor contributions which prevent the isolation of the input signals. A known offset, d(s), is added to the control input to remove the effects of these disturbances from the perspective of the controller, as illustrated in Fig. \ref{fig:controllercomps}a. 

The control inputs are defined as:  $\tau_\mathrm{u}\left(t\right) = d_{yaw}(t)\ +\ u_{yaw}(t)$ and $F_\mathrm{u}\left(t\right) = d_{alt}(t) + u_{alt}(t),$ where the disturbance terms are chosen to cancel out the coupling terms in Equation (\ref{eq:yawdynamics}) and Equation (\ref{eq:altdynamics}). The disturbance terms are then chosen as: 
\begin{equation}
\begin{gathered}
d_{\text {yaw }}(t)=I_{\text {rotor }} \cdot \ddot\alpha_{\text {rotor }}(t)+\frac{1}{2} \rho c_d A_{\text {ref }} r^3 \omega_{\text {rotor }}(t)^2, \\
d_{\text {alt }}(t)=m \cdot g-\frac{1}{2} \rho c_l A_{\text {ref }} r^2 \omega_{\text {rotor }}(t)^2 .
\end{gathered}
\end{equation}
In essence, the yaw disturbance $d_{yaw}(t)$ compensates for torque introduced from rotor acceleration and aerodynamic drag. The altitude disturbance $d_{alt}\left(t\right)$ compensates for the difference between the gravity force and the lift from the rotor.


\subsubsection{Effect of augmentation of plant transfer function}
In both yaw and altitude case, applying simplifying assumptions of rotor speed profile and a known offset generates a plant transfer function of
\begin{equation}\label{eq:closedlooptf}
    G_p(s)=\frac{G_v(s)}{s}=\frac{y(s)}{u(s)}=\frac{1}{\eta s^2},
\end{equation}

where $\eta_{\mathrm{yaw}}=I_{\mathrm{body}}$ and $\eta_{\mathrm{alt}}=m$. The transfer function outlined in Equation (\ref{eq:closedlooptf}) represents a marginally stable system, aligned with the analyses presented on Equation (\ref{eq:yawdynamics}) and Equation (\ref{eq:altdynamics}). 

\subsection{Stability Criteria}
The stability of each controller is evaluated by applying the Routh-Hurwitz criterion to the closed-loop transfer function that arises for the system shown in Fig. \ref{fig:controllercomps}. Under single-loop PID control, the controller is defined as $C_1\left(s\right)=k_{p,1}+k_{d,1}s+\frac{k_{i,1}}{s}$. Applying the Routh-Hurwitz criterion to the closed loop transfer function yields the requirement that ${k_{p,1}k}_{d,1}>\eta k_{i,1}$ for stable operation for yaw and altitude under both cases.
For the cascaded PI-PID controller, $C_1\left(s\right)=k_{p,1}\ +\ \frac{k_{i,1}}{s}$ and $C_2\left(s\right)=k_{p,2}+k_{d,2}s+\frac{k_{i,2}}{s}$. The derivative term is excluded from the outer loop due to the tendency to introduce excess noise. Applying the Routh-Hurwitz stability criterion to the closed loop transfer function of the system shown in Fig. \ref{fig:controllercomps}a produces the following nonlinear inequality involving all controller gains and system dynamics:
\begin{equation}\label{eq:cascadedstability}
\begin{aligned}[t]
&\left(\eta+k_{d2}\right)\left(k_{p1}k_{i2}+k_{i1}k_{p2}\right)^2
 + \left(k_{i1}k_{i2}\right)\left(k_{p2}+k_{p1}k_{d2}\right)^2  < \\
&\left(k_{p2}+k_{p1}k_{d2}\right)
   \left(k_{p1}k_{i2}+k_{i1}k_{p2}\right)
   \left(k_{i2}+k_{p1}k_{p2}+k_{i1}k_{d2}\right).
\end{aligned}
\end{equation}

\begin{figure}[!t]
    \centering
    \includegraphics[width=\linewidth]{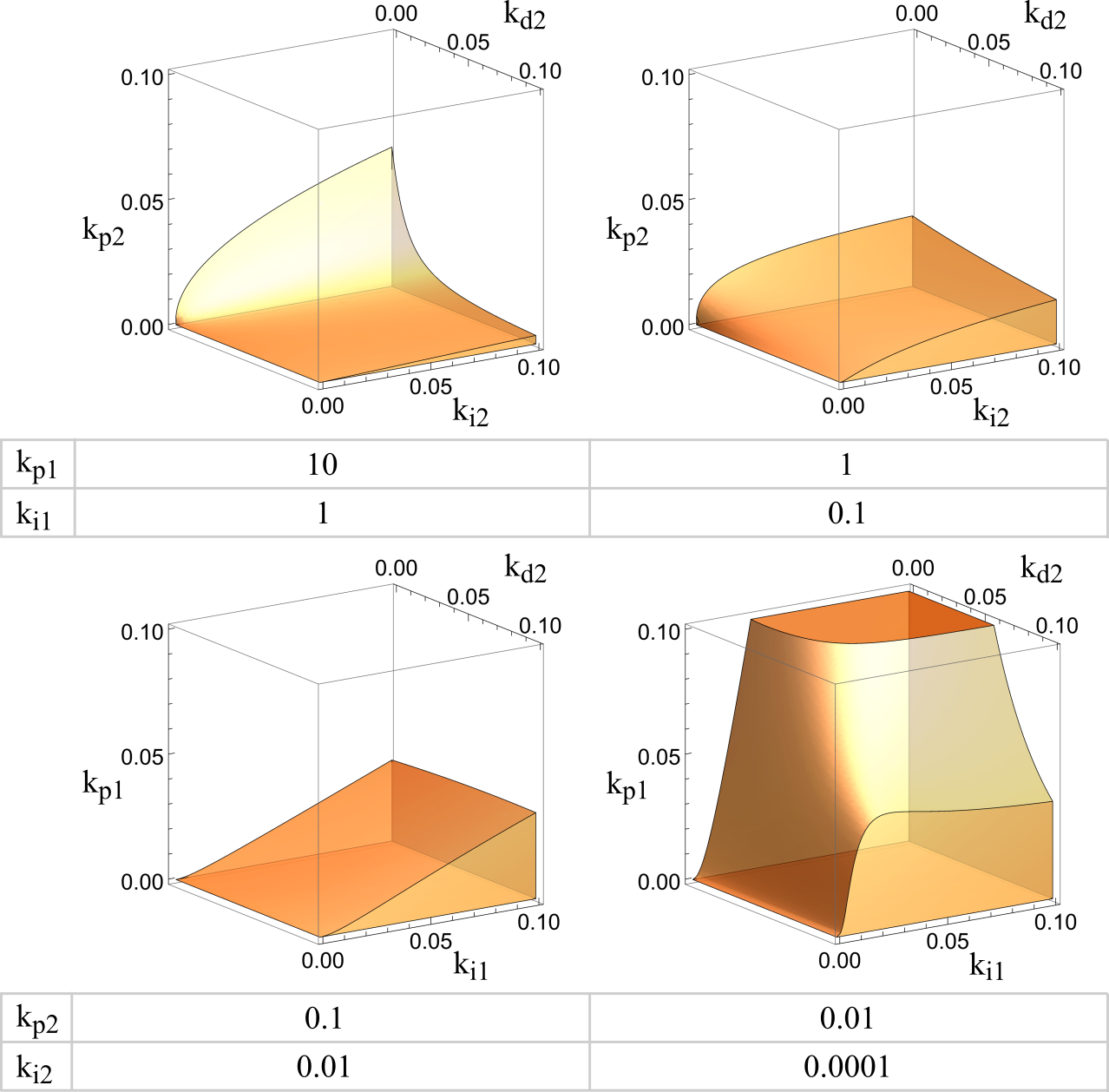}
    \caption{Yaw instability regions due to cascaded PID gains. The first row of graphs vary inner loop proportional and integral gains across panels. The second row of graphs vary outer loop proportional and integral gains across panels, while within each graph, the inner loop derivative gain and outer loop gains are varied. The orange regions of each plot indicate the gain combinations that produce unstable vehicle yaw.}
    \label{fig:instability_plots}
\end{figure}

Although the cascaded stability condition outlined in Equation (\ref{eq:cascadedstability}) is not directly suited for controller tuning, it can be used to verify the stability for a given set of controller gains and to qualitatively evaluate the relationship between gains. Fig. \ref{fig:instability_plots} presents instability region plots for differing values of proportional and integral gain of the inner and outer loops. Qualitatively, decreases in outer loop gains improves the stability margins of the system (Fig. \ref{fig:instability_plots}, top row), while decreases in inner loop proportional and integral gain decrease the stability margins of the system (Fig. \ref{fig:instability_plots}, bottom row).

\subsection{Simulated Controller Performance}
To evaluate the performance of the single-loop PID and cascaded PID controllers, we simulate the step response, disturbance rejection, modeling error rejection, and controller effort of each controller (Fig. \ref{fig:controllercomps}b--e). The gains used for performance comparison were determined via the optimization process outlined in Appendix \ref{app:pidoptimization}. The specific gain values are also reported in Appendix \ref{app:pidoptimization}.  

Both architectures exhibit a comparable step response (Fig. \ref{fig:controllercomps}b) and rejection of modeling error (Fig. \ref{fig:controllercomps}c), indicating similar tracking performance under nominal conditions. However, with a disturbance to the plant input (i.e., load disturbance), the cascaded controller asymptotically rejects the disturbance, whereas the single PID controller accumulates the error over time (Fig. \ref{fig:controllercomps}d), highlighting the improved robustness of the cascaded PID controller. In addition to improved load disturbance rejection, the cascaded PID controller reduces overall controller effort, particularly in peak demand (Fig. \ref{fig:controllercomps}e). These results support the cascaded controller as the preferred architecture for robust and energy-efficient yaw and altitude regulation on SPERO. 

\subsection{Key Operational Insights}
Preceding analyses from this section yield three main insights that directly inform SPERO's controller design and state-machine coordination: 
\begin{enumerate}
    \item Distinct forms of rotor speed ensures stability (i.e., Case I and Case II),
    \item Feedforward linearization reduces the effects of constant offset terms, 
    \item Cascaded PID provides closed form closed-loop stability margins,
\end{enumerate}

\section{State Machine Control for Stable Deployment}\label{sec:control}
The preceding sections identified the design features and control requirements necessary for the stable operation of a stopped-rotor aircraft, including maintaining lift continuity, managing auxiliary forces, and ensuring cross-regime stability. Building on these insights, this section describes how these requirements are integrated into SPERO’s operation through a mode-dependent controller implemented as a state machine.

The state machine framework coordinates three key elements: (1) activating the appropriate controller for each flight regime, (2) allocating auxiliary forces via the counterbalances, quadcopter subsystem, and center of pressure mechanism, and (3) dynamically reconfiguring vehicle geometry during transitions. In this way, the state machine operationalizes the principles derived in the preceding sections and enables robust mode switching across highly distinct flight regimes.

\begin{figure*}[!t]
    \centering
    \includegraphics[width=\textwidth]{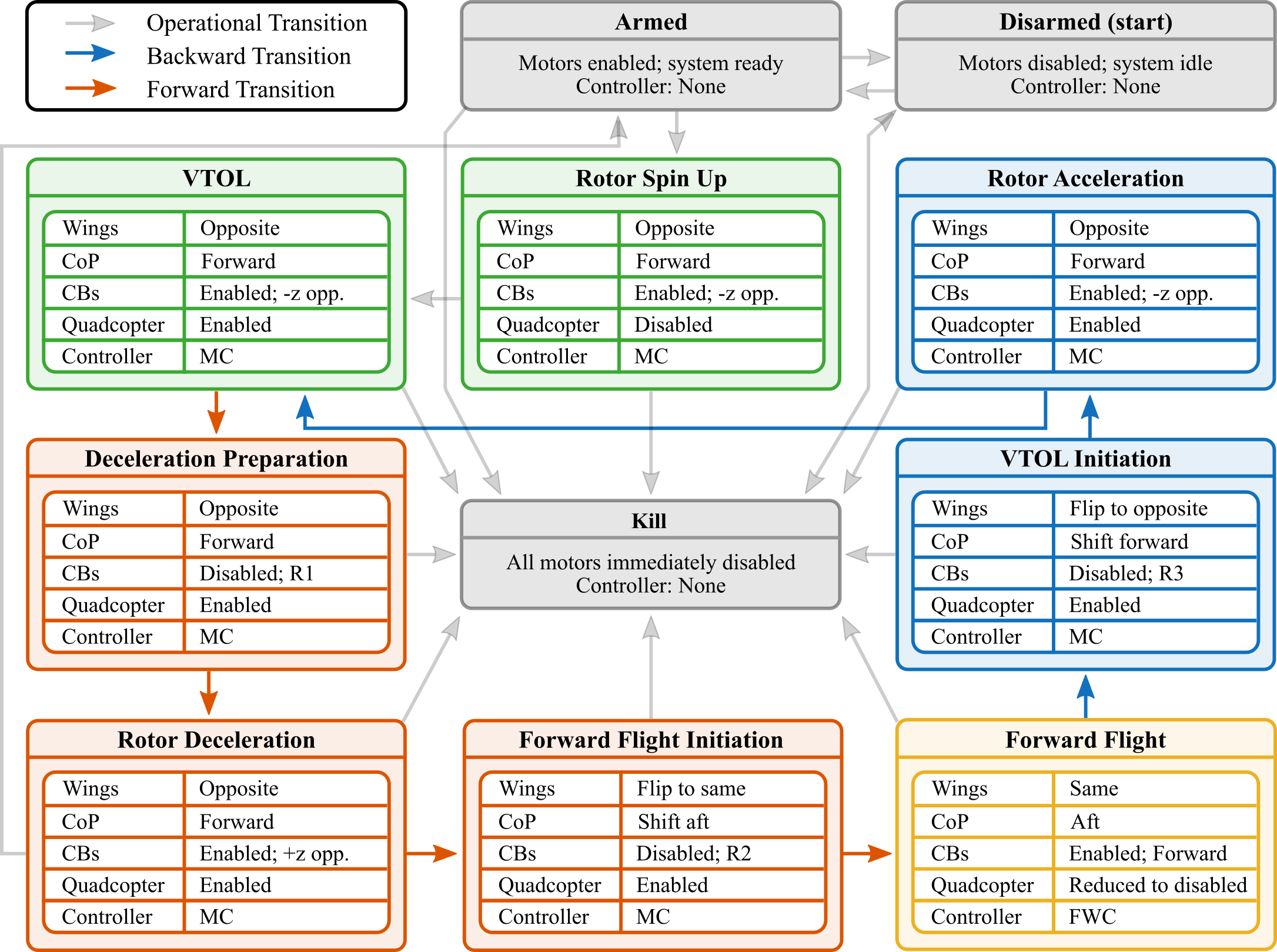}
    \caption{State machine diagram of SPERO where operational transitions necessary for safe operation are denoted by gray lines, forward transitions (i.e., from VTOL to forward flight) are shown with orange arrows, and backward transitions (i.e., from forward flight to VTOL) are denoted with blue arrows. Similarly, operational states are shown in gray, VTOL states are shown in green, forward transition states are shown in orange, forward flight is shown in yellow, and backward transition states are shown in blue. CoP refers to the position of the center of pressure, while CBs refer to the orientation of the counterbalances.}
    \label{fig:statemachine}
\end{figure*}

SPERO’s state machine comprises eleven discrete modes, grouped into five categories: safety, VTOL, forward flight, forward transition, and backward transition (Fig. \ref{fig:statemachine}). Each state uses one of two controller architectures: a multicopter (MC) controller for VTOL and transition phases, and a forward-wing controller (FWC) for fixed-wing flight. Both controllers are drawn from the PX4 open-source autopilot \cite{PX4Overview}, where they are standard, widely validated architectures used in conventional multicopter and fixed-wing vehicles. Integration with PX4 allows SPERO to leverage proven control modules while extending them through our custom state machine logic and transition sequencing. Full controller implementation details and transition conditions are provided in Appendix \ref{app:controller}.

\subsection{Safety States}
The safety states encompass the modes necessary to ensure smooth and secure operation of the vehicle, including a disarm, arm, and kill state, as shown by the gray shaded boxes in Fig. \ref{fig:statemachine}. The disarmed state is the starting state when the vehicle is powered on, but the motors are disabled (i.e., no signal) and the system is idle (i.e., computers and communication are active). In the armed state, the motors are enabled and commanded to zero speed and all systems are ready for flight. Finally, the kill state, which is reachable from every state of the vehicle, immediately disables all motors. These modes are used solely for operational handling of the vehicle and do not encompass any control policies.

\subsection{VTOL State}
From rest, VTOL is initiated with the rotor spin-up state, during which the vehicle remains grounded until the main rotor accelerates to a steady-state speed of 80 rad/s. Once the rotor reaches steady state, the system transitions to VTOL mode, where the vehicle enters a hover using lift from the main rotor. In VTOL mode, the vehicle is controlled in translational velocity, altitude, and yaw.




\subsection{Forward Flight States}
In the forward flight state, SPERO operates as a fixed-wing vehicle. The main rotor position is fixed, and lift is generated passively by air flow over the wings. As the vehicle transitions from rest (i.e., quadcopters generating all lift) to trimmed speed (i.e., fixed wings generating all lift), the controller used depends on the vehicle’s forward velocity. At lower cruise velocities ($<$ 10 m/s), the hybrid MC/FWC controller is used to ensure adequate altitude stability. At higher speeds ($>$ 10 m/s), the vehicle is controlled with a standard FWC, where the quadcopter is fully disabled and pitch, airspeed, and altitude are dynamically coupled.

\subsection{Forward Transition States}
Forward transition enables the reconfiguration of SPERO from VTOL to forward flight, following a sequence of three states depicted by orange arrows in Fig. \ref{fig:statemachine}. From the VTOL state, the vehicle enters the deceleration preparation state, where the counterbalances are temporarily disabled and the orientation is reversed. In this state, rotor speed is used to control yaw, while the quadcopter is used to control altitude.
 
Once the counterbalances have reversed direction, the vehicle enters the rotor deceleration state, where the main rotor is actively slowed at a constant rate. The counterbalances are used to counteract torque generated by the slowing rotor. Once the rotor is fully stopped and homed, the system enters the forward flight initiation state, during which the wing is flipped to the forward configuration, the counterbalances are rotated forward, and the center of pressure is shifted aft. When geometric configuration is complete, the vehicle transitions to the forward flight state.

\begin{figure*}[!b]
    \centering
    \includegraphics[width=0.95\textwidth]{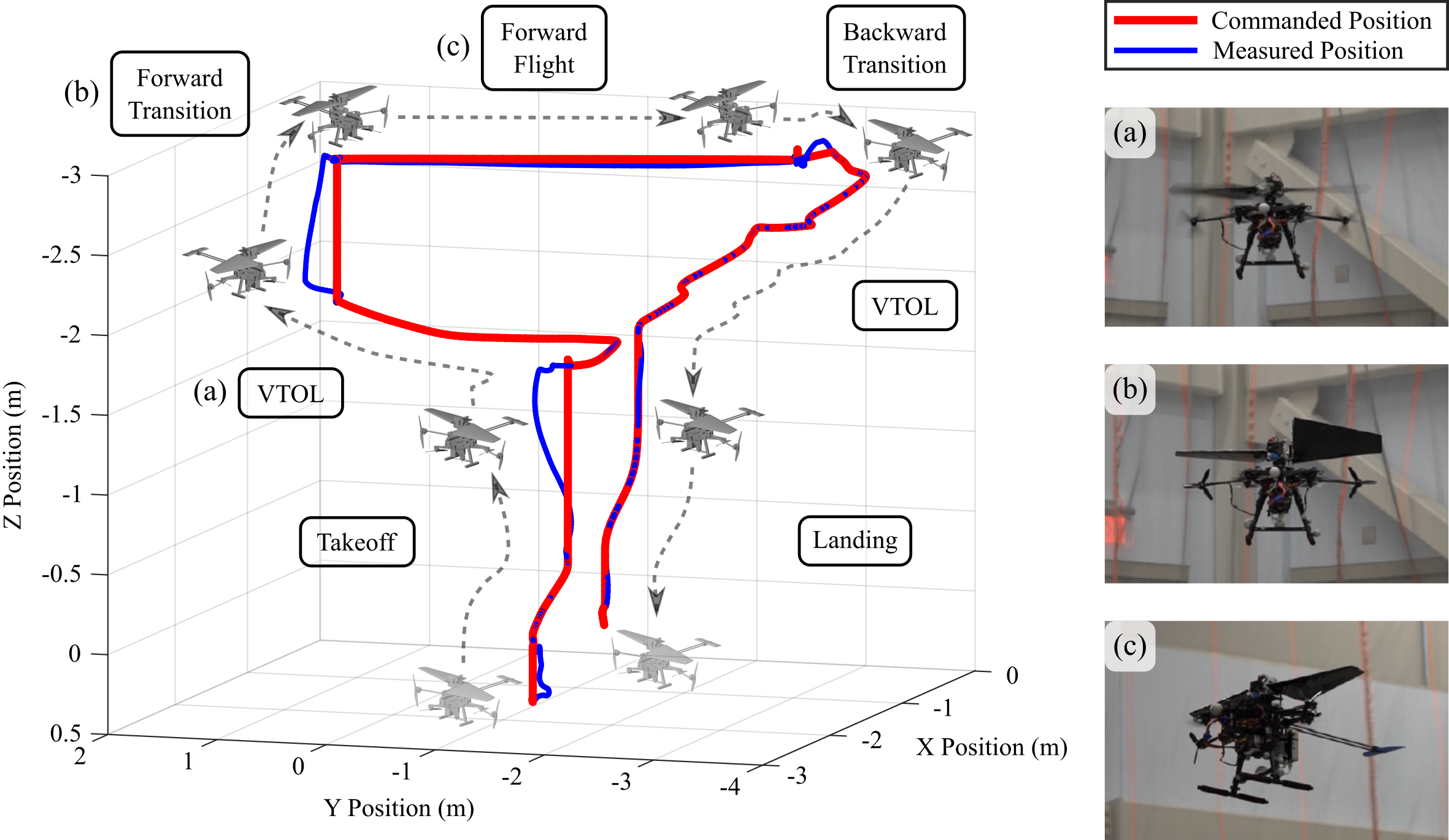}
    \caption{The 3-dimensional trajectory of SPERO during flight validation of bidirectional transition. Adjacent images show SPERO during the key phases of flight: (a) VTOL, (b) the wing flip maneuver during forward transition, and (c) forward flight. Measured position was obtained via a motion capture system, while velocity inputs were commanded by the user. In the absence of user input, the system maintained the current position.}
    \label{fig:flightvalidation_3d}
\end{figure*}

\subsection{Backward Transition States}
Backward transition returns the vehicle from forward flight to VTOL through two states (blue arrows, Fig. \ref{fig:statemachine}). When forward speed falls below 10 m/s, the vehicle enters VTOL initiation state. In this mode, the center of pressure is shifted forward, the counterbalances are rotated to prepare to counteract rotor acceleration, and the wing is flipped to the VTOL configuration. Once the geometric changes are complete, the vehicle enters the rotor acceleration state. Here, the rotor accelerates at a constant rate while the quadcopter holds the position, roll, and pitch of the vehicle. Yaw is controlled by the counterbalances. Upon completion, the system returns to the VTOL state.

\section{Experimental Demonstration by Bi-directional Transition}\label{sec:validation}
Real-world flight testing was performed to verify the stability of SPERO across flight modes and during transitions. An open-loop trajectory was commanded via a remote controller operated by a ground-based pilot. The commanded trajectory is depicted in red in Fig. \ref{fig:flightvalidation_3d}, and the actual trajectory is overlaid in blue. The complete flight, lasting approximately 120 s, is shown in entirety in the supplementary movie.

Flight validation was performed in a 12 m x 12 m x 7.5 m indoor motion capture facility outfitted with 20 Vicon cameras and a suspended safety net. SPERO was equipped with 12 retroreflective markers for full six-degree-of-freedom pose tracking.  Motion capture data was streamed at 100 Hz over WiFi to an onboard Raspberry Pi, where the data is transformed into the vehicle coordinate frame. The transformed data was transmitted to the PX4 flight controller over MAVLINK using MAVROS. The PX4 flight controller fuses the motion capture data into the integrated extended Kalman filter to update the state estimate.

For takeoff, SPERO was placed on a table in the motion capture space. All vehicle operations were initiated by a ground-based pilot using a standard RC transmitter. The pilot commanded body-relative x and y velocity, z position, and yaw rate. When x, y, and yaw sticks are in the neutral position, the vehicle enters a corresponding position hold mode. Desired trajectory is received onboard with a standard transmitter and communicated to the PX4 flight controller over SBUS. PX4 passes the desired setpoints to the low-level controller for the corresponding flight mode.

\subsection{Bidirectional Transition}
The bidirectional flight test included all key operational phases: SPERO took off and entered VTOL mode (Fig. \ref{fig:flightvalidation_3d}a). During forward transition, SPERO successfully performed the wing flip maneuver (Fig. \ref{fig:flightvalidation_3d}b), engaged the passive latch mechanism, and reoriented the center of pressure. The vehicle transitioned to forward flight and maintained stable operation across the length of the motion capture space (Fig. \ref{fig:flightvalidation_3d}c). Finally, SPERO performed a backward transition where the wing flipped back to the VTOL configuration, releasing the latch pin. The rotor accelerated to steady state. Once back in the VTOL state, the vehicle successfully landed. The entire flight is summarized in Fig. \ref{fig:flightvalidation_3d}.

This demonstration confirms that SPERO’s mechanical reconfiguration and control system enables stable operation through forward and backward transition modes. Trimmed forward flight could not be tested due to size limitations of the motion capture space. However, throughout VTOL and transition phases, no instability or loss of yaw or altitude control was observed, all factors that have contributed to the failure modes of prior vehicle designs.

\begin{figure}[!b]
    \centering
    \includegraphics[width=\linewidth]{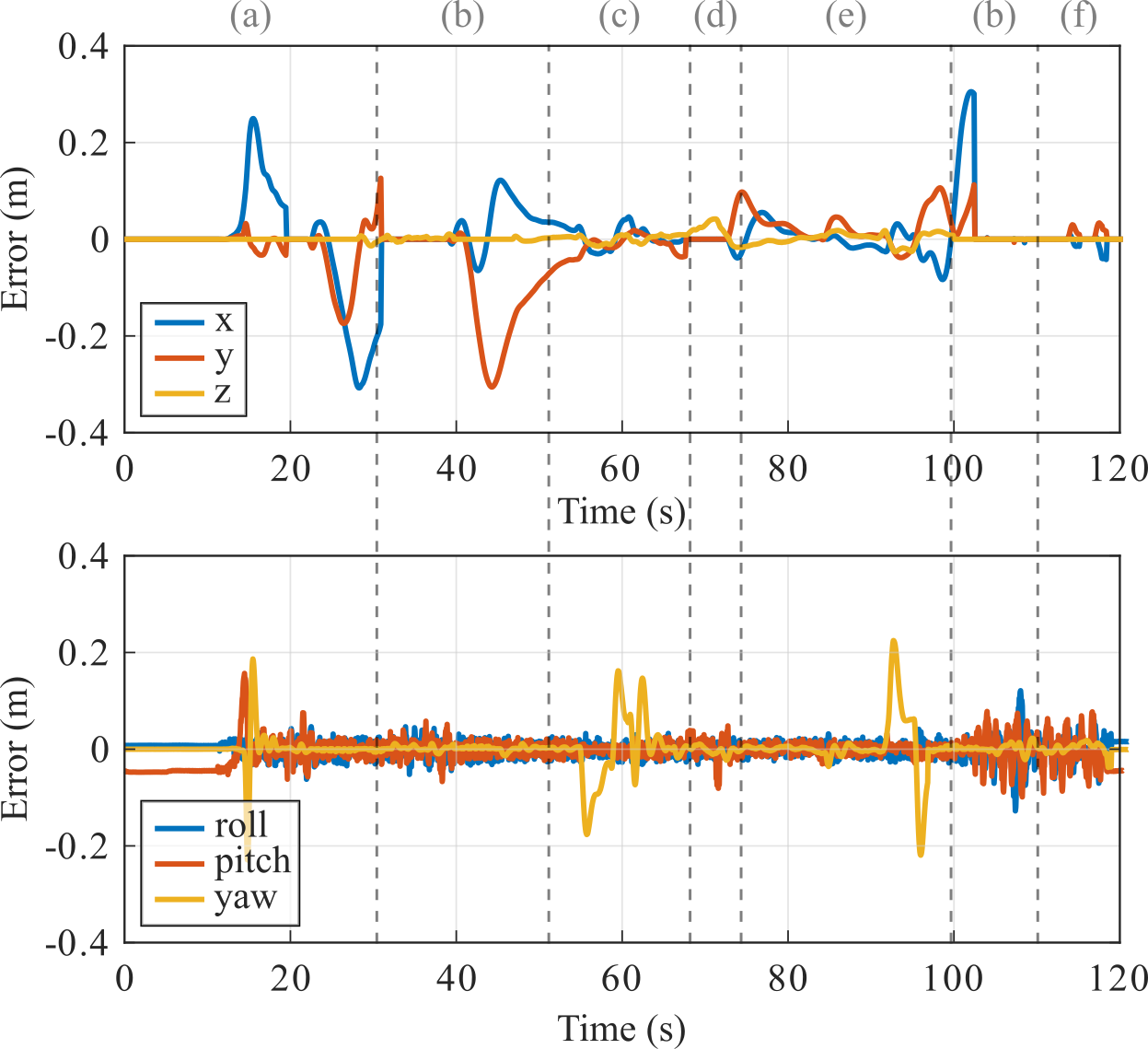}
    \caption{Time traces of the error between commanded and measured position and attitude during the bidirectional flight validation test. Dashed vertical gray lines denote the transitions between the different flight phases: (a) Takeoff, (b) VTOL, (c) forward transition, (d) forward flight, (e) backward transition, and (f) landing.}
    \label{fig:flightvalidation_error}
\end{figure}

\subsection{Closed-Loop Controller Tracking}
The error in commanded versus measured position and attitude across flight modes is summarized in Fig. \ref{fig:flightvalidation_error}. The largest deviations occurred during takeoff (Fig. \ref{fig:flightvalidation_error}a) and VTOL mode (Fig. \ref{fig:flightvalidation_error}b), with root mean square error (RMSE) of 0.074 m in $x$, 0.064 m in $y$, and 0.008 m in $z$. The initial jump in error during takeoff corresponds to a brief physical interaction of SPERO’s landing gear with the net, as shown in the supplementary movie. Subsequent jumps in x and y position error correlate to commanded yaw inputs and diminish sharply once the yaw command is removed. This behavior suggests a misalignment between SPERO’s center of gravity and the center of thrust of the counterbalances (Fig. \ref{fig:reducedmodel}), causing lateral drifts during rotation commands.

The RMSE for roll, pitch, and yaw, are 0.017 rad, 0.027 rad, and 0.040 rad, respectively. Yaw experiences the largest deviations, with spikes occurring in takeoff (Fig. \ref{fig:flightvalidation_error}a), forward transition (Fig. \ref{fig:flightvalidation_error}c), and backward transition (Fig. \ref{fig:flightvalidation_error}e). The disturbance during forward and backward transition is correlated with dynamic maneuvers, including wing flipping and impulse acceleration of the rotors; however, the disturbances attenuate once the action is complete.  Roll and pitch tracking remained consistent throughout the flight test.

Together, these results indicate that SPERO maintains stable tracking despite the reconfiguration of aerodynamic surfaces, redistribution of control forces during transition, and the introduction of impulse forces. The absence of oscillations or divergence in the error graphs (Fig. \ref{fig:flightvalidation_error}) following transition suggests that the onboard mechanical and control systems effectively stabilize SPERO.

\section{Conclusion}\label{sec:conclusion}
In this work, we introduce SPERO, a 2.7 kg stopped-penta rotor UAV that achieves the first experimentally validated, stable, and reversible bidirectional transition between VTOL and forward flight. In addition to demonstrating feasibility, we identify four feasibility conditions (FC1 -- FC4) that define the aerodynamic, geometric, auxiliary force, and lift-continuity requirements for practical stopped-rotor designs, establishing a general framework for future vehicles. From these insights, we design an integrated design and control methodology that combines a flipping and latching wing, an active center of pressure mechanism, thrust vectored counterbalances, and a multicopter system, all of which are coordinated by an 11 state machine. By pairing simplified dynamic models with controller evaluation, we identify the conditions necessary to guarantee yaw stability and altitude regulation during highly nonlinear stopped-rotor transitions.

While the design and experiments confirm the effectiveness of SPERO, new challenges are exposed. First, energy efficiency remains a limitation during hover and transitions, suggesting opportunities for design optimization. Additionally, outdoor flight testing in unconstrained environments is needed to fully characterize the performance. Nonetheless, the results presented here lay design and control groundwork for a new class of efficient VTOL UAV. 



{\appendices

\begin{table*}[!t]
\centering
\caption{State transition table for SPERO’s 11-state mode-dependent controller. Inputs: \textbf{Kill} = manual safety override, \textbf{Arm} = vehicle arming command, \textbf{State} = commanded flight mode. State names match those in Fig. \ref{fig:statemachine}.}
\label{tab:state_transitions}
\renewcommand{\arraystretch}{1.2}
\begin{tabular}{c|cccc|c}
\hline
\textbf{Current State} &
\multicolumn{4}{c|}{\textbf{Inputs}} &
\textbf{Next State} \\
\cline{2-5}
 & \textbf{Kill} & \textbf{Arm} & \textbf{State Command} & \textbf{Other} & \\
\hline
Disarmed                  & 0 & 1   & 0 (Null)           &   --                          & Armed \\
Armed                     & 0 & 0   & 0 (Null)           &   --                          & Disarmed \\
                          & 0 & 1   & 1 (VTOL)           &   --                          & Rotor Spin Up \\
Rotor Spin Up             & 0 & 1   & 1 (VTOL)           &   Rotor Speed = 0             & VTOL \\
VTOL                      & 0 & 1   & 2 (Forward Flight) &   --                          & Deceleration Preparation \\
Deceleration Preparation  & 0 & 1   & 2 (Forward Flight) &   Counterbalance Speed = 0    & Rotor Deceleration \\
Rotor Deceleration        & 0 & 1   & 2 (Forward Flight) &   Rotor Speed = 0             & Forward Flight Initiation \\
Forward Flight Initiation & 0 & 1   & 2 (Forward Flight) &   Counterbalance Speed = 0    & Forward Flight \\
Forward Flight            & 0 & 1   & 2 (Forward Flight) &   Vehicle Speed = 0           & VTOL Initiation \\
VTOL Initiation           & 0 & 1   & 1 (VTOL)           &   Counterbalance Speed = 0    & Rotor Acceleration \\
Rotor Acceleration        & 0 & 1   & 1 (VTOL)           &   Rotor Acceleration = 0      & VTOL \\
Kill                      & 0 & 1   & 0 (Null)           &   --                          & Disarmed \\
\hline
Any                       & 1 & 0/1 & Any                &   --                          & Kill \\
\hline
\end{tabular}
\end{table*}

\begin{table*}[!b]
\centering
\caption{PID controller gains used on SPERO}
\label{table:controller_gains}
\begin{tabular}{l l c c c l l c}
\hline
\textbf{Controller} & \textbf{Type} & \textbf{P} & \textbf{I} & \textbf{D} & \textbf{Saturation} & \textbf{Units} & \textbf{Integral Windup} \\
\hline
\multicolumn{8}{c}{\textit{Multi-copter (MC)}} \\
\cline{1-8}
Roll              & Attitude & 6.500 & 0.000 & 0.000 & $[-3.840,\ 3.840]$ & rad/s & N/A \\
Roll Rate         & Rate     & 0.150 & 0.200 & 0.003 & $[-1.000,\ 1.000]$ & rad/s & $[-0.300,\ 0.300]$ \\
Pitch             & Attitude & 6.500 & 0.000 & 0.000 & $[-3.840,\ 3.840]$ & rad/s & N/A \\
Pitch Rate        & Rate     & 0.150 & 0.200 & 0.003 & $[-1.000,\ 1.000]$ & rad/s & $[-0.300,\ 0.300]$ \\
Yaw               & Attitude & 2.800 & 0.000 & 0.000 & $[-3.840,\ 3.840]$ & rad/s & N/A \\
Yaw Rate          & Rate     & 0.200 & 0.100 & 0.000 & $[-1.000,\ 1.000]$ & rad/s & $[-0.300,\ 0.300]$ \\
X and Y Position  & Position & 0.950 & 0.000 & 0.000 & N/A                & m     & N/A \\
X and Y Velocity  & Velocity & 1.800 & 0.400 & 0.200 & $[-12.000,\ 12.000]$ & m/s  & N/A \\
Z Position        & Position & 1.000 & 0.000 & 0.000 & $[-1.500,\ 3.000]$ & m     & N/A \\
Z Velocity        & Velocity & 4.000 & 2.000 & 0.000 & N/A                & m/s   & N/A \\
\hline
\multicolumn{8}{c}{\textit{Forward Flight (FWC)}} \\
\cline{1-8}
Roll              & Attitude & 2.500 & 0.000 & 0.000 & $[-1.221,\ 1.221]$ & rad/s & N/A \\
Roll Rate         & Rate     & 0.050 & 0.100 & 0.000 & N/A                & rad/s & $[-0.200,\ 0.200]$ \\
Pitch             & Attitude & 2.500 & 0.000 & 0.000 & $[-1.047,\ 1.047]$ & rad/s & N/A \\
Pitch Rate        & Rate     & 0.080 & 0.100 & 0.000 & N/A                & rad/s & $[-0.400,\ 0.400]$ \\
Yaw               & Attitude & 2.500 & 0.000 & 0.000 & $[-0.873,\ 0.873]$ & rad/s & N/A \\
Yaw Rate          & Rate     & 0.050 & 0.100 & 0.000 & N/A                & rad/s & $[-0.200,\ 0.200]$ \\
Energy Rate       & Rate     & 0.050 & 0.020 & 0.000 & N/A                & N/A   & N/A \\
Energy Balance    & Control  & 0.100 & 0.100 & 0.000 & N/A                & N/A   & N/A \\
\hline
\end{tabular}
\end{table*}

\section{Maximum Center of Pressure to Center of Gravity Calculation}\label{app:copcalc}
The movement of the center of gravity relative to the center of gravity of the body is computed with the following formula:
\begin{equation}\label{eq:cgcp}
C_{g, x}=\frac{m_{\text {rail }} r_{\text {rail }}+m_{\text {wing }} r_{\text {wing }}}{m},
\end{equation}
where $m=2.7$ kg is the total mass of the vehicle, $ m_{\mathrm{wing}}\ =\ 0.34$ kg is the mass of both wings, $ m_{\mathrm{rail}}\ =\ 0.51$ kg is the mass of the components that move along the linear rail, $r_{\mathrm{rail}}$  is the distance between the body center of mass and the center of gravity of the rail assembly, and $r_{\mathrm{wing}}$ is the distance between the body center of mass and the center of gravity of the wings. 

In the VTOL configuration, $r_{\mathrm{rail}} = r_{\mathrm{wing}} = 0$, as the system is designed to axially align these bodies through the axis of rotation of the vehicle. Evaluating Equation (\ref{eq:cgcp}) for VTOL yields $C_{g,x}\ =\ 0$ m, consistent with the alignment of the rail, wing, and main body through the axis of rotation of the vehicle.

In forward flight, the center of gravity inevitably shifts as the wing rotates and the rail assembly translates. In SPERO $r_{\mathrm{rail}}\ =\ 0.05$ m corresponds to the maximum displacement of the active center of pressure, and $r_{\mathrm{wing}}\ =\ 0.08$ m. Substituting these values into Equation (\ref{eq:cgcp}) yields $C_{g,x}\ =\ 0.02$ m. Given the linear servo stroke of 0.05 m, the resulting maximum offset between the shifted center of pressure and center of gravity is approximately 0.03 m, as reported in Table \ref{table:designparams} in the main text. 

\section{PID Parameter Optimization}\label{app:pidoptimization}
PID gains for comparing single-loop and cascaded PID control were determined by formulating a constrained optimization program. For all optimizations, the \texttt{MATLAB} \texttt{fmincon} function was used to minimize a scalar cost function of the form:
\begin{equation}
    J = \mathrm{IAE} + \lambda \cdot \int_{0}^{T} u^2(t) \, dt,
\end{equation}
where $\mathrm{IAE}$ is the integral of absolute error between the reference and the measured output over a 1-second time horizon, and $\lambda$ is a regularization parameter used to penalize control effort $u(t)$. The initial guess for all gains was unity, while the parameters were constrained to be positive and less than $100$.  

The resulting single-loop gains were determined to be $k_{p1} = 0.004$, $k_{i1} = 0.010$, and $k_{d1} = 0.561$. The cascaded gains were determined to be $k_{p1} = 13.100$, $k_{i1} = 0.002$, $k_{p2} = 13.600$, $k_{i2} = 0.036$, and $k_{d2} = 1.370 \times 10^{-5}$.


\section{Controller Architecture}\label{app:controller}
The quadcopter of SPERO utilizes a hybrid control system implemented within the PX4 software, combining the multicopter and fixed-wing controllers \cite{PX4Overview}. During VTOL and hover, the vehicle utilizes the standard PX4 cascaded PID multicopter controller, referred to as MC in Fig. \ref{fig:statemachine}. In forward flight, the vehicle utilizes the standard PX4 fixed-wing controller, denoted FWC in Fig. \ref{fig:statemachine}. Transitions between the MC and FWC controllers are handled internally by PX4. The reader is directed to the PX4 documentation \cite{PX4Overview} for more information on the implementation of and transition between these controllers.

The transition conditions for the 11 state machine are summarized in Table \ref{tab:state_transitions}. The PID controller gains across flight modes are summarized in Table \ref{table:controller_gains}. 
	

}


\bibliographystyle{IEEEtran}
\bibliography{references.bib}

\vfill

\end{document}